\documentclass[10pt,twocolumn,letterpaper]{article}

\usepackage[pagenumbers]{wacv} 

\newcommand{\blue}[1]{{\color{black}#1}}

\newcommand{\commentout}[1]{}
\usepackage{algpseudocode}
\usepackage[ruled,vlined]{algorithm2e}
\usepackage{multirow}
\usepackage{booktabs}   
\usepackage{makecell}   

\definecolor{wacvblue}{rgb}{0.21,0.49,0.74}
\usepackage[pagebackref,breaklinks,colorlinks,allcolors=wacvblue]{hyperref}

\def\wacvPaperID{1434} 
\def\confName{WACV}
\def\confYear{2027}

\title{SettleSplat: Monocular Gaussian Mapping under In-View Scene Changes}

\author{
Anh Thuan Tran\textsuperscript{1,2,*} \quad
Xiyun Song\textsuperscript{2,$^{\dagger}$} \quad
Yubin Zhou\textsuperscript{2} \quad
Heather Yu\textsuperscript{2} \quad
Jana Kosecka\textsuperscript{1} \quad
Liang Peng\textsuperscript{2}\\[4pt]
\textsuperscript{1}George Mason University \qquad
\textsuperscript{2}Futurewei Technologies\\[2pt]
{\tt\small \{atran92, kosecka\}@gmu.edu \quad \{xsong, yzhou2, hyu, lpeng\}@futurewei.com}
}

\begin{document}
\twocolumn[{%
\renewcommand\twocolumn[1][]{#1}%
\maketitle

\begin{center}
    \centering
    \includegraphics[width=1\textwidth]{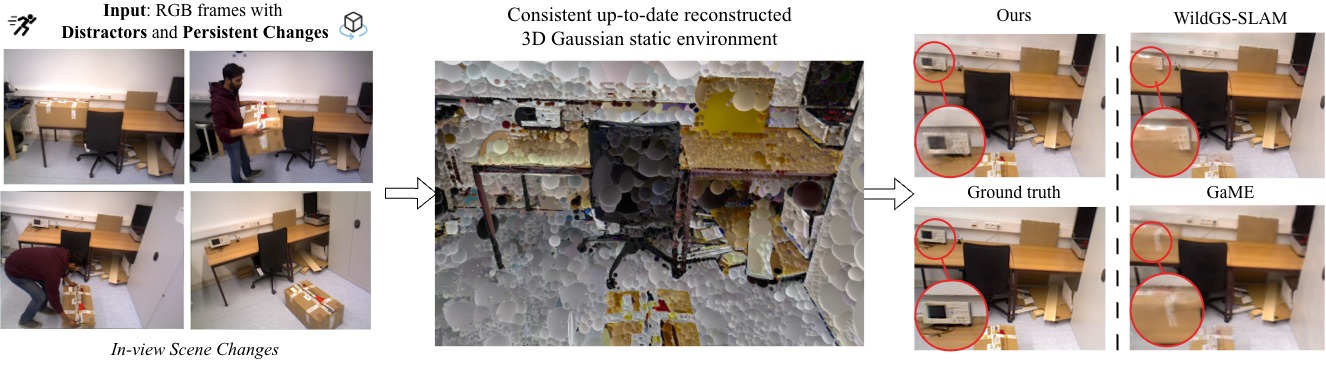}
    \captionof{figure}{\textbf{SettleSplat.} Given monocular RGB sequence (left) with \textit{in-view scene changes}, containing dynamic distractors (moving person) and persistent changes (box), SettleSplat reconstructs a consistent, up-to-date 3D Gaussian map (middle). At previously changed locations (red circles), our method achieves clean and consistent rendering, while previous studies ~\cite{zheng2025wildgs, yugay2026game} leave ghosting artifacts (right).}
    \label{fig:teaser}
\end{center}%

}]

\renewcommand*{\thefootnote}{\fnsymbol{footnote}}
\footnotetext[1]{Work done during an internship at Futurewei Technologies.}
\footnotetext[2]{Corresponding author.}

\begin{abstract}
Online 3D Gaussian mapping enables fast and high-fidelity novel view synthesis, with broad applications in augmented reality, robotics, and autonomous systems. 
However, many current methods assume static scenes or filter out dynamics as outliers. 
In this paper, we study in-view scene changes where objects are manipulated in front of the camera within a single online sequence, unlike prior works that detect changes outside the camera view or between separate captures.
This setting entangles transient and persistent dynamics, as objects can be distractors while moving and persistent once they settle.
While recent studies start to address distractors by removing them for clean static background, in-view scene changes remain underexplored. 
To resolve this, we present SettleSplat, a monocular mapping system that maintains up-to-date 3D Gaussian maps under such changes.
Specifically, we lift coarse per-pixel uncertainty into Gaussian dynamicity field, then prune transient ones. 
For persistent changes, we derive geometric contradictions weighted by parameter-free rendered depth uncertainty to separate true changes from under-optimized regions. 
On Bonn and Co-Fusion, SettleSplat achieves consistent 3D Gaussian maps, outperforming current baselines on both full images and change regions while maintaining comparable speed.
\end{abstract}


\section{Introduction}
\label{sec:intro}
Visual mapping aims to reconstruct an up-to-date and consistent 3D representation from RGB images. It is a fundamental task in computer vision, with broad applications in robotics, autonomous systems, and AR/VR \cite{fei20243dsurvey}, where spatial understanding supports navigation, planning, and interaction. 
Recent developments in 3D representations, such as NeRF \cite{mildenhall2021nerf} or 3D Gaussian Splatting \cite{kerbl3Dgaussians}, further provide novel view synthesis capabilities with photorealistic rendering. Although achieving remarkable performance in both image quality and training speed, these approaches \cite{keetha2024splatam, tran2026varsplat, sandstrom2025splatslam} assume that the scene remains static during optimization across all frames.

However, real-world scenes are more complicated with different kinds of motions. 
For example, a person walking through the scene is treated as dynamic distractor, while an object newly placed into the scene is a persistent change that must be reflected in the final representation.
These two dynamics, however, are intertwined within a single sequence in the camera's view, a setting we refer to as \textit{in-view scene changes}.
When a hand places an item into the scene, the body movement is considered a distractor, while the item, once settled, becomes persistent change. Prior works address either distractor removal \cite{zheng2025wildgs,li2026droid, bescos2018dynaslam, li2025garadslam},  change detection between separate captures \cite{lu20253dgscd, ackermann2025cl}, or changes that happen out of view in a single sequence \cite{yugay2026game}. The mixed case, where objects transition from dynamic to static or vice versa within camera view, remains unresolved.
The monocular setting makes this problem harder due to errors in depth estimation \cite{yang2024depthanything2, piccinelli2024unidepth, hu2024metric3d}. Previous studies \cite{zheng2025wildgs,li2026droid} show that DINOv2-based uncertainty, derived from semantic features can correct these errors in bundle adjustment for tracking. For mapping, however, the uncertainty map is too coarse, and estimated depth alone is not enough to capture persistent changes from Gaussians that are still under-optimized and keep changing during online reconstruction.

To address these challenges, we propose SettleSplat, a monocular method for\commentout{in-view scene changes} change detection where dynamic distractors and persistent changes are entangled.
For distractors, we lift the coarse 2D uncertainty into per-Gaussian dynamicity field and prune dynamic ones from the 3D map. For persistent changes, we capture geometric contradictions between the map and observed frames, weighted by a rendered depth uncertainty that separates persistent changes from under-optimized Gaussians. 
To the best of our knowledge, SettleSplat is the first monocular 3DGS mapping system that handles persistent changes entangled with transient distractors  without depth sensor or segmentation masks. 
In summary, our contributions are as follows.

\begin{itemize}
    \item We propose SettleSplat,\commentout{an online monocular mapping method} a monocular method for online mapping of dynamic scenes where transient distractors and persistent changes are entangled. 
    \item We introduce a per-Gaussian dynamicity field that lifts coarse 2D uncertainty into 3D for pruning, and a correction mechanism for recovery of wrongly suppressed regions.
    \item We derive a rendered depth uncertainty \commentout{without learnable parameters, then use it }to gate geometric contradictions between the map and previous frames, separating persistent changes from under-optimized regions. 
    \item Our method outperforms current approaches on in-view scene changes, containing dynamic distractors and persistent changes at competitive tracking and mapping speed.
\end{itemize}

\section{Related Work}
\label{sec:related}
\textbf{Dynamic Gaussian Splatting.}
Neural radiance fields \cite{mildenhall2021nerf} enable novel view synthesis with photorealistic rendering. 
3D Gaussian Splatting \cite{kerbl3Dgaussians} further improves rendering speed and optimization time, achieving real-time performance. 
Extending this to dynamic scenes has received broad attention. 
Existing methods model temporal changes through deformation fields \cite{yang2024deformable,wu20244dgs} or per-Gaussian motion representations \cite{luiten2024dynamicgaussians,lin2024gaussianflow,kratimenos2024dynmf, lei2025mosca}, enabling dynamic novel view synthesis. However, these approaches typically rely on offline reconstruction and multi-view or RGB-D observations, whereas SettleSplat performs online mapping from monocular RGB sequences.
\\
\textbf{Online Gaussian Mapping.}
Early online mapping systems use neural implicit representations, encoding the scene into MLP or feature grid that is optimized on-the-go for novel view synthesis \cite{sucar2021imap, zhu2024nicerslam, sandstrom2023pointslam}. 
Explicit 3DGS then accelerated this line, replacing volumetric sampling with rasterized Gaussians for high-fidelity rendering at faster speed. 
Recent studies use 3D Gaussians as the main scene representation for tracking and mapping \cite{keetha2024splatam,monogs,yugay2023gaussianslam,peng2024rtgslam}.
Subsequent works extend the setting along several axes, such as reconstructing from monocular RGB \cite{monogs, huang2024photoslam, sandstrom2025splatslam}, modeling uncertainty to down-weight unreliable observations \cite{tran2026varsplat,hu2024cgslam}, and using loop closure to remain consistent over long trajectories \cite{zhu2025loopsplat,liso2024loopyslam}, then typically completing the online map with a short final refinement. 
However, these works compute uncertainties against observed color \cite{tran2026varsplat} or sensor depth \cite{hu2024cgslam}, whereas we render depth uncertainty purely from Gaussian geometry to mark reliable regions.
\\
\textbf{Distractor Removal.}
While these methods perform well in static scenes, they still struggle in dynamic environments. Another line of works therefore removes dynamic distractors through semantic or motion masks \cite{bescos2018dynaslam, jiang2024rodyn, li2025garadslam} or transient-aware modeling in the wild \cite{martin2021nerfw, li2026droid}. WildGS-SLAM \cite{zheng2025wildgs} learns an uncertainty map from DINOv2 features with a shallow MLP \cite{kulhanek2024wildgaussians}, while DROID-W \cite{li2026droid} obtains uncertainty from DINOv2 feature consistency inside bundle adjustment. However, these approaches treat dynamics in image space and only remove the distractors, not considering regions that transition from dynamic to persistent. We use DROID-W only as an uncertainty source and lift its per-pixel uncertainty into 3D Gaussian dynamicity field, deferring the masking decision to 3D instead of removing pixels in image space.
\begin{figure*}[t]
    \centering
    \includegraphics[width=\textwidth]{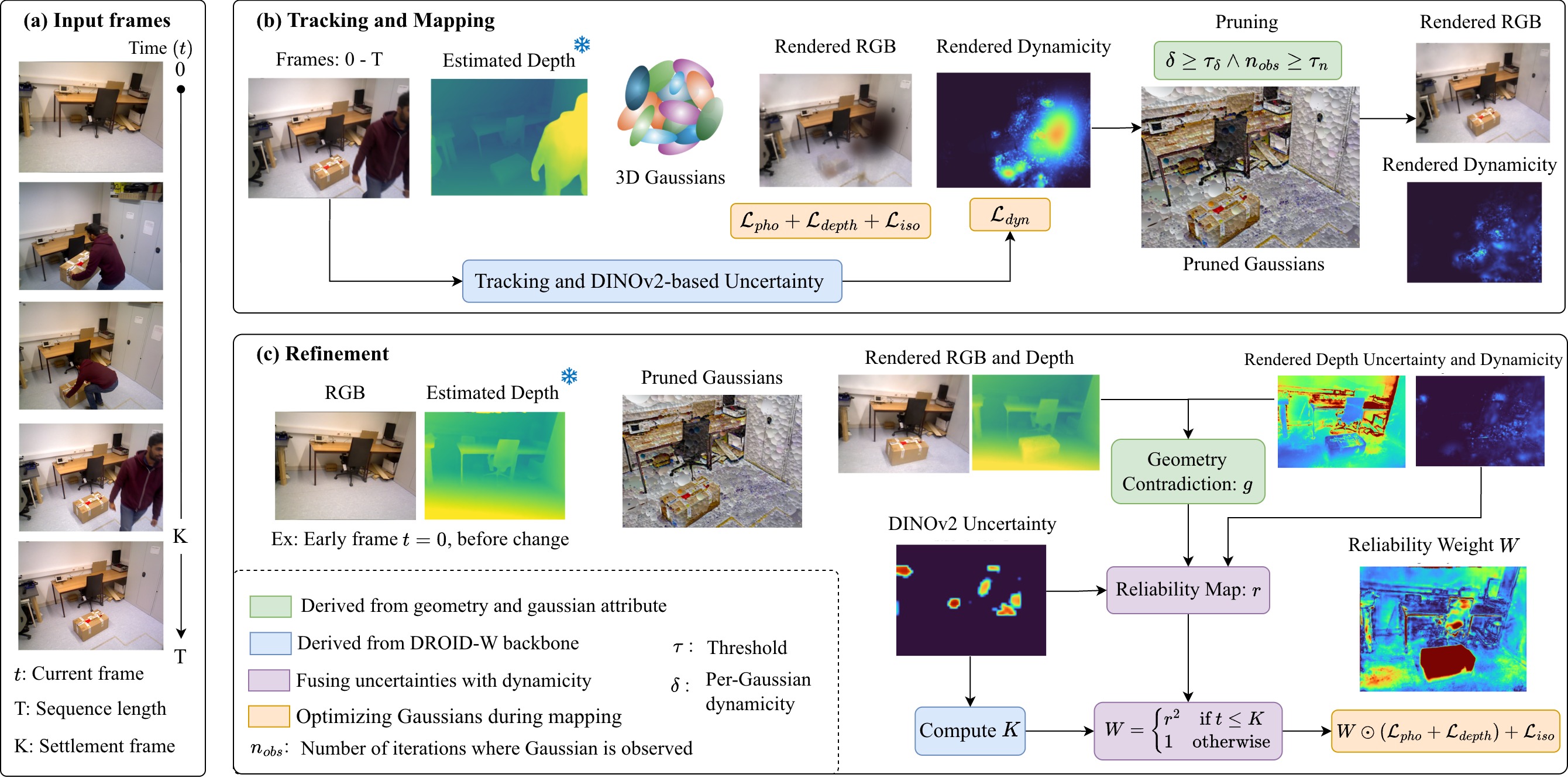}
    \caption{\textbf{System overview.} Given monocular RGB frames (a), SettleSplat estimates depth, then tracks and maps with the DROID-W backbone (b) while learning per-Gaussian dynamicity field $\delta$ that prunes transient distractors. During refinement (c), geometry contradiction score $g$, derived from the rendered map and estimated depth and weighted by the rendered depth uncertainty, is integrated with the tracking uncertainty into reliability map $r$. Frames before the settlement frame $K$ are weighted by $W=r^2$, while later frames supervise the settled scene without masking.}
    \label{fig:pipeline}
\end{figure*}
\\
\textbf{3DGS Change Detection.}
To address persistent changes and maintain up-to-date 3DGS map, several studies have been proposed \cite{lu20253dgscd, kim2025towards} to detect change regions and update the scene from sparse post-change images without re-optimizing the whole map. Another line of work \cite{ackermann2025cl,zeng2025gaussianupdate,kim2026ltgs} applies continual learning to 3DGS, maintaining the map across separate captures and lighting conditions using DINOv2 features and clustering algorithms. 
\blue{Unlike these approaches, which update Gaussian maps from sparse observations across different scene states, SettleSplat handles transient and persistent changes intertwined within a continuous monocular sequence.}
GaME \cite{yugay2026game} addresses this issue by bringing map updating into a single online sequence, focusing on objects that are changed outside the camera view. However, it requires posed RGB-D input and instance segmentation from pretrained models \cite{ravi2025sam2}. In contrast, our method addresses \commentout{in-view scene} changes that unfold in front of the camera, where distractors and persistent changes are entangled in the same sequence. SettleSplat separates these motions, removing distractors while updating the map, and operates from monocular RGB without segmentation mask.

\section{Method}
SettleSplat adapts uncertainty-aware SLAM backbone DROID-W~\cite{li2026droid} to in-view scene changes with mapping stage that reconstructs up-to-date 3D Gaussian maps.
Given monocular RGB sequences where transient distractors and persistent changes are entangled, our system removes the former while updating the map to reflect the latter.
The overview of SettleSplat is shown in Figure 2.

\subsection{Preliminaries}
\textbf{3D Gaussian Splatting.}
Following \cite{kerbl3Dgaussians,keetha2024splatam,zheng2025wildgs,li2026droid},
we represent the scene as a set of 3D Gaussians, each parameterized by mean
position $\boldsymbol{\mu}_i$, covariance $\boldsymbol{\Sigma}_i$ (factored
into scale $s_i$ and rotation $q_i$), opacity $\alpha_i$, and color
$\mathbf{c}_i$ represented by spherical harmonics. In addition, each Gaussian
has a learnable dynamicity logit $\ell_i$, mapped through a sigmoid to
obtain a bounded dynamicity value $\delta_i$:
\begin{equation}
    \mathcal{G}_i = \{\boldsymbol{\mu}_i, \mathbf{c}_i,
    s_i, q_i, \alpha_i, \ell_i\},
    \quad \delta_i = \sigma(\ell_i)\in(0,1).
\end{equation}
Rendering follows standard alpha compositing: with transmittance $T_i$, the
blending weight of each Gaussian is $w_i = T_i\alpha_i$, and color and depth
are accumulated as
\begin{equation}
    \hat C=\sum_i w_i\mathbf c_i, \qquad \hat D=\sum_i w_i d_i.
\end{equation}
\textbf{Dynamicity rendering.}
Unlike color or depth, dynamicity $\delta_i$ is a surface attribute rather than an emitted quantity. At pixels with low accumulated opacity, plain accumulation would conflate coverage with dynamicity. We therefore normalize by the accumulated blending weights to obtain $\hat M$,
\begin{equation}
    \hat M= \frac{\sum_i w_i\delta_i}{\sum_i w_i + \varepsilon},
\end{equation}
giving an opacity-independent estimate $\hat M\in(0,1)$. Color, depth, and
dynamicity are thus rendered for every pixel in a single rasterization pass.
\\
\textbf{Uncertainty-aware backbone.}
We build on DROID-W \cite{li2026droid}, which extends DROID-SLAM \cite{teed2021droid} with uncertainty-aware bundle adjustment. While tracking, DROID-W provides a per-pixel uncertainty map, derived from multi-view consistency of DINOv2 features \cite{oquab2023dinov2} across views. Higher values indicate observations that violate the rigid-scene assumption, such as moving objects. We adapt DROID-W tracking stage and use its uncertainty $p_{\mathrm{ba}}\in(0,1)$ as supervision for the dynamicity loss (Sec. 3.3). Our contributions lie in the mapping stage. While DROID-W handles tracking in the presence of moving objects, we keep the 3DGS map consistent when the scene
changes.

\subsection{Geometry contradiction} 
\textbf{Depth changes.} Transient distractors can be pruned from the map, but persistent changes cannot. When an object moves to a new place, pruning alone leaves a ghost of its old position in the map. To capture such changes, we compare the map against early frames. We render depth from the up-to-date map and compute its residual against the estimated depth of previous frames. 
Unlike GaME \cite{yugay2026game}, which requires depth sensor, we only use
monocular depth $D$ estimated by a pretrained model \cite{hu2024metric3d}, similar to \cite{zheng2025wildgs, li2026droid}. Large residual here does not always mean a persistent change. It can also come from under-optimized Gaussians, especially in growing regions of the map.
\\
\textbf{Depth uncertainty.}  To separate the sources of large residual apart, we derive a rendered depth uncertainty directly from Gaussian geometry, without additional learnable parameters. Each Gaussian encodes depth variation through its spatial covariance. We first project its 3D covariance onto the camera axis:
\begin{equation}
    \sigma^2_{d,i} = \mathbf{e}_z^{\top}\mathbf{W}_t \boldsymbol{\Sigma}_i \mathbf{W}_t^{\top}\mathbf{e}_z,
\end{equation}
where $\mathbf{W}_t$ is the world-to-camera rotation at frame $t$ and
$\mathbf{e}_z=[0,0,1]$ is the unit vector along the camera axis. 
\blue{This projection is consistent with the z-depth used in rendering and depth comparison, but differs from ray-conditioned depth variance.}
Aggregating these per-splat depth variances follows the second-moment alpha compositing in VarSplat \cite{tran2026varsplat}. Using the same blending weights $w_i = T_i \alpha_i$, we obtain per-pixel variance.
\begin{equation}
    V_d = \sum_i w_i\!\left(\sigma^2_{d,i} + d_i^2\right) - \hat{D}^2.
\end{equation}
Unlike VarSplat \cite{tran2026varsplat}, which learns per-splat appearance variance, our depth variance $\sigma_{d,i}^2$ is determined analytically from the projected 3D covariance, where $d_i$ denotes the Gaussian's camera-space depth. 
Therefore, per-pixel depth variance $V_d$ can be rendered in forward-pass rasterization.

We then use $V_d$ to weight the depth residual from estimated depth $D$ and rendered depth $\hat D$. Pixels with high uncertainty are discounted, so attributing the large residuals \commentout{are attributed }to persistent changes rather than to under-optimized Gaussians.
This enables us to compute the geometry contradiction map $g$. 
\begin{equation}
    g = \operatorname{clamp}\!\Bigg(\frac{|D-\hat{D}|}{V_d+\epsilon},\,0,\,1\Bigg).
\end{equation}
We clamp the ratio rather than normalize it, since per-frame normalization would rescale $g$ by each frame's residual range. A fixed clamp keeps $g$ comparable frame to frame.

\subsection{Online mapping}
\textbf{Lifting 2D uncertainty to 3D Gaussians.} To build a clean static map, prior works~\cite{zheng2025wildgs,li2026droid} mask or downweight dynamic pixels directly in the 2D input images. This strategy has two problems. First, 2D masks cover not only moving objects but also the static background behind them, so background Gaussians are wrongly suppressed. 
Second, it mishandles objects that transition from dynamic to static. When such an object settles, the mask still marks it dynamic and suppresses Gaussians that now belong to the static scene.
This gets worse with coarse uncertainty maps~\cite{li2026droid}, which spill beyond moving object into nearby static regions. We take the opposite approach by mapping every frame without masking and deferring decision to 3D. Each Gaussian carries a dynamicity logit supervised by coarse per-pixel uncertainty from tracking, and once mapping is done, dynamic Gaussians are removed by thresholds on dynamicity and observation count. 
\\
\textbf{Mapping losses.} Concretely, we create and optimize Gaussians on all pixels, following
common practice \cite{zheng2025wildgs,li2026droid},
\begin{equation}
    \mathcal{L}_\text{pho} = (1-\lambda_\text{SSIM})\,|\hat C - C|
    + \lambda_\text{SSIM}\,(1-\mathrm{SSIM}),
\end{equation}
\begin{equation}
    \mathcal{L}_\text{depth} = |\hat D - D|, \qquad
    \mathcal{L}_\text{dyn} = \mathrm{BCE}\big(\hat M,\ p_\text{ba}\big).
\end{equation}
The dynamicity loss is a binary cross-entropy (BCE) between the rendered dynamicity $\hat M$ and the DROID-W uncertainty $p_{ba}$ as the target. This uncertainty is RGB-based, derived from DINOv2 feature consistency across views, and is distinct from the geometric depth uncertainty $V_d$ in Section 3.2.
The rendering pipeline itself lifts this 2D map into 3D. 
Occluded Gaussians have low transmittance, so their normalized weights $w/\sum_i w_i$ are smaller and get less gradient. Therefore, the supervision mainly updates the dynamicity of Gaussians on the moving object in front. 
The full online loss combines the reconstruction part with the dynamicity and isotropic regularizer that controls the scale of Gaussians \cite{monogs,li2026droid},
\begin{equation}
    \mathcal{L}_\text{online}
    = \lambda_\text{pho}\mathcal{L}_\text{pho}
    + (1-\lambda_\text{pho})\mathcal{L}_\text{depth}
    + \lambda_\text{dyn}\,\mathcal{L}_\text{dyn}
    + \lambda_\text{iso}\, \lVert\hat s- s\rVert_1.
\end{equation}
\blue{Following \cite{yugay2023gaussianslam, zhu2025loopsplat}, the isotropic regularizer penalizes differences between Gaussian scales and their mean to discourage elongated Gaussians.}
\\
\textbf{Dynamic pruning.}
After mapping optimization, we prune Gaussians with consistently high dynamicity,
\begin{equation}
    \text{prune } \mathcal{G}_i \iff \big(\delta_i > \tau_{\delta}\big)
    \ \wedge\ \big(n_i^{\text{obs}} > \tau_{n}\big),
\end{equation}
where $n_i^{\text{obs}}$ counts the mapping iterations in which $\mathcal{G}_i$ is observed. The observation threshold prevents pruning Gaussians due to false alarms, preserving scene coverage.


\subsection{Refinement}

Following standard baselines \cite{yugay2023gaussianslam, yugay2026game, monogs, zheng2025wildgs, li2026droid}, SettleSplat ends with refinement over the whole sequence. However, instead of treating every frame equally, our refinement is mainly driven by a per-pixel reliability map $r$ that captures changed regions. 
\\
\textbf{Reliability map.}
During refinement, moving distractors and objects in pre-change regions can confuse the Gaussian optimization. To address this, we build a per-pixel reliability map $r$ that captures both distractors and persistent changes. The static part $(1-p_\text{ba})$ is trusted by default. Pixels flagged as dynamic by $p_\text{ba}$ are partially recovered through the rendered dynamicity $\hat M_\text{pruned}$ after pruning. If a pixel still renders high dynamicity, the pruning removed nothing there, so the flag is likely a false alarm due to the coarse uncertainty map. Finally, we multiply by $(1-g)$ so that pixels with geometric contradictions remain downweighted,
\begin{equation}
    r = \big[\, (1-p_{\mathrm{ba}}) + \hat M_\text{pruned} \odot p_{\mathrm{ba}} \big]\,(1 - g).
\end{equation}
Therefore, once an object is fully mapped at its new location, $g$ vanishes and $r$ recovers that region automatically. 
\blue{This mechanism allows the approach to accommodate multiple scene changes, such as sequential placements, while the settlement frame $K$ provides a single global transition for final refinement.}
\\
\textbf{Settlement frame $K$.} In the transition tail near the end of a sequence, the tracking uncertainty $p_{ba}$ does not drop immediately. It keeps marking a just-settled object as dynamic for several frames because $p_\text{ba}$ is driven by multi-view feature inconsistency \cite{li2026droid}, which persists for a while after the object stops moving. Left to $r$ alone, these frames are still suppressed and the object is under-optimized. To handle this, we use settlement frame $K$ to remove this lag, letting the map catch up to the change. 
Specifically, we compute how much the tracking uncertainty $p_{ba}$ changes between consecutive frames. A settlement is a frame-level attribute rather than a local one, so we average this change over all pixels $\Omega$ of that frame into a single activity value, and smooth it with a moving average of window $\omega$ so that one noisy frame cannot trigger the threshold on its own,
\begin{equation}
    a_t = \tfrac{1}{|\Omega|}\sum_{p\in\Omega} \big|\,p_{ba}(t) - p_{ba}(t-1)\,\big|,
     \tilde a_t = \mathrm{MA}_\omega(a_t)
\end{equation}
\blue{We use tracking uncertainty as a proxy for scene activity, which is expected to decrease as the scene settles.} We take the last time $\tilde a_t$ exceeds a threshold $\theta = \tau_\text{settle}\cdot \max_t \tilde a_t$, relative to the peak activity, and set $K$ to the following frame. For distractor-only sequences where the objects never settle, $K$ falls near the end of the sequence.

\begin{equation}
    K = t_{\,[\,\max\{i:\ \tilde a_i > \theta\}\,+\,1\,]}.
\end{equation}
From equations 11 and 13, frames up to $K$ are weighted by $r^2$, while later frames supervise the map with all pixels,

\begin{equation}
    W = \begin{cases} r^2 & t \le K \\ 1 & t > K\end{cases}
\end{equation}
The refinement loss then weights each pixel by $W$ and keeps the isotropic regularizer,
\begin{equation}
    \mathcal{L}_\text{refine} = W \odot \big(\lambda_\text{pho}\mathcal{L}_\text{pho}
    + (1-\lambda_\text{pho})\mathcal{L}_\text{depth} \big) + \lambda_\text{iso}\, \lVert\hat s- s\rVert_1.
\end{equation}


\section{Experiment}
\textbf{Datasets.}
We evaluate on the Bonn RGB-D Dynamic dataset \cite{bonn} and Co-Fusion \cite{cofusion}. 
From each dataset, we use all sequences involving in-view object changes, whether objects are added, moved, or removed. This gives 7 sequences on Bonn and 4 on Co-Fusion, covering all three change types.
In-view changes remain scarce in existing benchmarks, which mostly capture distractor-only motion.
For each sequence, we annotate a reference frame only to build the held-out test split. The exact frames are listed in the supplementary. Starting from this frame, we hold out every 5th frame on Bonn and every 3rd frame on Co-Fusion, whose sequences are shorter, as novel test views. The remaining frames are the input sequence for all methods. Note that the method itself does not assume a single change. It computes $K$ from the input sequence, without access to the held-out views or the annotation. All methods are evaluated on the same held-out views for fair comparison.
\\
\textbf{Metrics.} Following standard evaluation~\cite{kerbl3Dgaussians,zheng2025wildgs,monogs}, we report PSNR, SSIM, and LPIPS between rendered and ground-truth images at full resolution. 
In addition, we manually annotate change regions in each sequence with bounding boxes from ground truth images. 
The regions cover where the scene changes, including old and new locations of moved objects, newly placed objects, and locations where objects are removed.
We crop the rendered and ground-truth images with the same boxes for all methods and compute rendering metrics on the cropped regions. More details are provided in supplementary. 
\\
\textbf{Baselines.} We compare SettleSplat with SplaTAM~\cite{keetha2024splatam}, MonoGS~\cite{monogs}, WildGS-SLAM~\cite{zheng2025wildgs}, DROID-W~\cite{li2026droid}, and GaME~\cite{yugay2026game} on the same sequences and held-out views. These baselines differ in how much scene dynamics they support. SplaTAM and MonoGS assume a static scene, while WildGS-SLAM and DROID-W remove dynamic distractors. GaME further handles persistent changes, but requires posed RGB-D input and instance masks. As RGB-D methods, GaME and SplaTAM are provided sensor depth, and GaME additionally uses generated instance masks \cite{ravi2025sam2}. SettleSplat, WildGS-SLAM, and DROID-W are evaluated from monocular RGB with estimated depth \cite{hu2024metric3d}. All hyperparameters of our method are listed in supplementary and are identical across all 11 sequences on both datasets, with no per-scene tuning.

\begin{table*}[t]
\centering
\caption{Novel view synthesis results on the Bonn dynamic dataset. We report PSNR, SSIM, and LPIPS on the full image (top) and within the change region (bottom). \textbf{Best} and \underline{second-best} results are highlighted.}
\label{tab:bonn}
\resizebox{\textwidth}{!}{%
\begin{tabular}{l l cccccccc}
\toprule
Method & Metric & \texttt{moving} & \texttt{moving2} & \texttt{placing} & \texttt{placing2} & \texttt{placing3} & \texttt{removing} & \texttt{removing2} & Avg. \\
\midrule
\multicolumn{10}{c}{\textit{Full Image}} \\
\midrule
\multirow{3}{*}{SplaTAM \cite{keetha2024splatam}} & PSNR$\uparrow$ & 21.38 & 22.84 & 20.51 & 18.74 & 22.75 & 21.95 & 19.39 & 21.08 \\
 & SSIM$\uparrow$ & 0.817 & 0.847 & 0.771 & 0.778 & 0.847 & 0.851 & 0.824 & 0.819 \\
 & LPIPS$\downarrow$ & 0.267 & 0.254 & 0.306 & 0.351 & 0.244 & 0.235 & 0.291 & 0.278 \\
\cmidrule(lr){1-10}
\multirow{3}{*}{MonoGS~\cite{monogs}} & PSNR$\uparrow$ & 18.89 & 19.88 & 17.09 & 21.64 & 20.91 & \textbf{24.01} & 19.37 & 20.26 \\
 & SSIM$\uparrow$ & 0.696 & 0.745 & 0.640 & 0.794 & 0.777 & \underline{0.852} & 0.790 & 0.756 \\
 & LPIPS$\downarrow$ & 0.416 & 0.398 & 0.501 & 0.339 & 0.303 & 0.259 & 0.340 & 0.365 \\
\cmidrule(lr){1-10}
\multirow{3}{*}{WildGS-SLAM \cite{zheng2025wildgs}} & PSNR$\uparrow$ & \underline{25.37} & 23.11 & 20.63 & 20.38 & 22.78 & \underline{23.83} & \textbf{22.14} & 22.61 \\
 & SSIM$\uparrow$ & \underline{0.880} & 0.842 & 0.829 & 0.814 & 0.832 & \textbf{0.856} & 0.805 & 0.837 \\
 & LPIPS$\downarrow$ & \underline{0.175} & 0.296 & 0.286 & 0.336 & 0.274 & \underline{0.235} & 0.368 & 0.281 \\
\cmidrule(lr){1-10}
\multirow{3}{*}{GaME \cite{yugay2026game}} & PSNR$\uparrow$ & 24.58 & \textbf{25.42} & \textbf{22.32} & 23.27 & \underline{24.15} & 23.72 & 20.52 & \underline{23.42} \\
 & SSIM$\uparrow$ & 0.807 & 0.834 & 0.795 & 0.799 & 0.805 & 0.810 & 0.776 & 0.804 \\
 & LPIPS$\downarrow$ & 0.199 & 0.204 & 0.251 & 0.265 & 0.238 & \textbf{0.228} & 0.308 & 0.242 \\
\cmidrule(lr){1-10}
\multirow{3}{*}{DROID-W \cite{li2026droid}} & PSNR$\uparrow$ & 24.74 & 24.88 & 21.87 & \textbf{23.54} & 23.66 & 22.86 & 21.02 & 23.22 \\
 & SSIM$\uparrow$ & 0.875 & \underline{0.889} & \underline{0.846} & \underline{0.882} & \underline{0.876} & 0.841 & \underline{0.833} & \underline{0.863} \\
 & LPIPS$\downarrow$ & 0.189 & \underline{0.189} & \underline{0.242} & \underline{0.193} & \underline{0.202} & 0.259 & \underline{0.288} & \underline{0.223} \\
\cmidrule(lr){1-10}
\multirow{3}{*}{\textbf{SettleSplat (Ours)}} & PSNR$\uparrow$ & \textbf{25.51} & \underline{25.39} & \underline{22.23} & \underline{23.43} & \textbf{24.76} & 22.77 & \underline{21.69} & \textbf{23.68} \\
 & SSIM$\uparrow$ & \textbf{0.884} & \textbf{0.899} & \textbf{0.849} & \textbf{0.886} & \textbf{0.887} & 0.840 & \textbf{0.846} & \textbf{0.870} \\
 & LPIPS$\downarrow$ & \textbf{0.167} & \textbf{0.165} & \textbf{0.235} & \textbf{0.190} & \textbf{0.173} & 0.254 & \textbf{0.247} & \textbf{0.205} \\
\midrule
\multicolumn{10}{c}{\textit{Change Region}} \\
\midrule
\multirow{3}{*}{SplaTAM \cite{keetha2024splatam}} & PSNR$\uparrow$ & 19.18 & 19.63 & 18.84 & 14.35 & 19.11 & 20.73 & 17.47 & 18.47 \\
 & SSIM$\uparrow$ & 0.783 & 0.765 & 0.611 & 0.546 & 0.656 & 0.800 & 0.794 & 0.708 \\
 & LPIPS$\downarrow$ & 0.265 & 0.257 & 0.413 & 0.638 & 0.431 & 0.362 & 0.392 & 0.394 \\
\cmidrule(lr){1-10}
\multirow{3}{*}{MonoGS~\cite{monogs}} & PSNR$\uparrow$ & 16.33 & 17.44 & 15.28 & 17.95 & 18.95 & \textbf{24.29} & 16.76 & 18.14 \\
 & SSIM$\uparrow$ & 0.640 & 0.643 & 0.471 & 0.611 & 0.638 & 0.811 & 0.757 & 0.653 \\
 & LPIPS$\downarrow$ & 0.442 & 0.434 & 0.596 & 0.415 & 0.313 & 0.395 & 0.494 & 0.441 \\
\cmidrule(lr){1-10}
\multirow{3}{*}{WildGS-SLAM \cite{zheng2025wildgs}} & PSNR$\uparrow$ & \underline{23.87} & 22.48 & 15.26 & 12.57 & 16.52 & 22.86 & \textbf{24.09} & 19.66 \\
 & SSIM$\uparrow$ & \underline{0.878} & 0.812 & 0.656 & 0.612 & 0.665 & 0.811 & 0.840 & 0.753 \\
 & LPIPS$\downarrow$ & \underline{0.146} & 0.238 & 0.514 & 0.615 & 0.540 & 0.319 & 0.338 & 0.387 \\
\cmidrule(lr){1-10}
\multirow{3}{*}{GaME \cite{yugay2026game}} & PSNR$\uparrow$ & 20.96 & 21.40 & 19.24 & 19.52 & \underline{20.77} & \underline{23.93} & 17.12 & 20.42 \\
 & SSIM$\uparrow$ & 0.763 & 0.762 & 0.668 & 0.672 & 0.683 & 0.821 & 0.765 & 0.733 \\
 & LPIPS$\downarrow$ & 0.203 & 0.192 & 0.349 & 0.250 & \underline{0.213} & 0.297 & 0.444 & 0.278 \\
\cmidrule(lr){1-10}
\multirow{3}{*}{DROID-W \cite{li2026droid}} & PSNR$\uparrow$ & 23.14 & \underline{22.70} & \underline{19.69} & \underline{22.13} & 19.72 & 22.62 & 20.06 & \underline{21.44} \\
 & SSIM$\uparrow$ & 0.863 & \underline{0.864} & \underline{0.784} & \underline{0.844} & \underline{0.806} & \underline{0.851} & \underline{0.853} & \underline{0.838} \\
 & LPIPS$\downarrow$ & 0.178 & \underline{0.142} & \underline{0.253} & \underline{0.148} & 0.295 & \underline{0.249} & \underline{0.300} & \underline{0.224} \\
\cmidrule(lr){1-10}
\multirow{3}{*}{\textbf{SettleSplat (Ours)}} & PSNR$\uparrow$ & \textbf{25.59} & \textbf{24.80} & \textbf{21.24} & \textbf{23.84} & \textbf{25.75} & 22.83 & \underline{22.40} & \textbf{23.78} \\
 & SSIM$\uparrow$ & \textbf{0.880} & \textbf{0.886} & \textbf{0.828} & \textbf{0.886} & \textbf{0.883} & \textbf{0.857} & \textbf{0.874} & \textbf{0.871} \\
 & LPIPS$\downarrow$ & \textbf{0.141} & \textbf{0.100} & \textbf{0.170} & \textbf{0.094} & \textbf{0.101} & \textbf{0.229} & \textbf{0.210} & \textbf{0.149} \\
\bottomrule
\end{tabular}}
\end{table*}

\subsection{Experimental Results}

\begin{table*}[t]
\centering
\caption{Novel view synthesis results on the CoFusion dataset, evaluated on the full image and within the change region. \textbf{Best} and \underline{second-best} results are highlighted.}
\label{tab:cofusion}
\resizebox{\textwidth}{!}{%
\begin{tabular}{l l ccccc ccccc}
\toprule
 & & \multicolumn{5}{c}{Full Image} & \multicolumn{5}{c}{Change Region} \\
\cmidrule(lr){3-7} \cmidrule(lr){8-12}
Method & Metric & \texttt{place} & \texttt{sliding} & \texttt{handover} & \texttt{car} & Avg. & \texttt{place} & \texttt{sliding} & \texttt{handover} & \texttt{car} & Avg. \\
\midrule
\multirow{3}{*}{SplaTAM~\cite{keetha2024splatam}} & PSNR$\uparrow$ & 22.42 & 29.08 & 13.91 & 28.98 & 23.60 & 20.89 & 31.59 & 13.85 & 16.99 & 20.83 \\
 & SSIM$\uparrow$ & 0.844 & 0.848 & 0.292 & 0.955 & 0.735 & 0.791 & 0.876 & 0.225 & 0.614 & 0.626 \\
 & LPIPS$\downarrow$ & 0.249 & 0.223 & 0.770 & 0.099 & 0.335 & 0.286 & \underline{0.322} & 0.806 & 0.307 & 0.430 \\
\cmidrule(lr){1-12}
\multirow{3}{*}{MonoGS~\cite{monogs}} & PSNR$\uparrow$ & 15.71 & 28.57 & 20.53 & 32.14 & 24.24 & 14.16 & 29.53 & 20.59 & 20.90 & 21.29 \\
 & SSIM$\uparrow$ & 0.713 & 0.807 & 0.750 & 0.967 & 0.809 & 0.614 & 0.868 & 0.703 & 0.715 & 0.725 \\
 & LPIPS$\downarrow$ & 0.350 & 0.275 & 0.277 & 0.058 & 0.240 & 0.470 & 0.428 & 0.290 & 0.354 & 0.386 \\
\cmidrule(lr){1-12}
\multirow{3}{*}{WildGS-SLAM~\cite{zheng2025wildgs}} & PSNR$\uparrow$ & \underline{24.77} & 28.75 & 19.68 & \underline{35.71} & 27.23 & \underline{23.36} & \textbf{36.16} & 20.60 & \underline{28.11} & \underline{27.06} \\
 & SSIM$\uparrow$ & \underline{0.909} & 0.797 & 0.748 & 0.979 & 0.858 & \underline{0.883} & \textbf{0.893} & 0.694 & \underline{0.924} & 0.848 \\
 & LPIPS$\downarrow$ & \underline{0.183} & 0.238 & 0.468 & \underline{0.033} & 0.231 & \underline{0.207} & 0.326 & 0.530 & \underline{0.050} & 0.278 \\
\cmidrule(lr){1-12}
\multirow{3}{*}{GaME~\cite{yugay2026game}} & PSNR$\uparrow$ & 23.02 & 19.10 & 19.94 & 22.46 & 21.13 & 17.37 & 17.63 & 26.48 & 25.14 & 21.65 \\
 & SSIM$\uparrow$ & 0.905 & 0.750 & 0.711 & 0.799 & 0.791 & 0.623 & 0.677 & 0.753 & 0.790 & 0.711 \\
 & LPIPS$\downarrow$ & 0.254 & 0.496 & 0.226 & 0.155 & 0.283 & 0.525 & 0.537 & 0.468 & 0.157 & 0.422 \\
\cmidrule(lr){1-12}
\multirow{3}{*}{DROID-W~\cite{li2026droid}} & PSNR$\uparrow$ & 20.71 & \underline{30.75} & \textbf{29.66} & 33.68 & \underline{28.70} & 19.18 & 34.06 & \textbf{29.60} & 19.85 & 25.67 \\
 & SSIM$\uparrow$ & 0.850 & \underline{0.855} & \textbf{0.916} & \underline{0.983} & \underline{0.901} & 0.800 & 0.880 & \textbf{0.905} & 0.851 & \underline{0.859} \\
 & LPIPS$\downarrow$ & 0.271 & \underline{0.198} & \textbf{0.104} & 0.037 & \underline{0.153} & 0.331 & 0.348 & \textbf{0.106} & 0.293 & \underline{0.269} \\
\cmidrule(lr){1-12}
\multirow{3}{*}{\textbf{SettleSplat (Ours)}} & PSNR$\uparrow$ & \textbf{36.56} & \textbf{31.03} & \underline{28.72} & \textbf{39.05} & \textbf{33.84} & \textbf{35.75} & \underline{35.11} & \underline{28.29} & \textbf{33.36} & \textbf{33.12} \\
 & SSIM$\uparrow$ & \textbf{0.956} & \textbf{0.855} & \underline{0.900} & \textbf{0.984} & \textbf{0.924} & \textbf{0.950} & \underline{0.893} & \underline{0.885} & \textbf{0.965} & \textbf{0.923} \\
 & LPIPS$\downarrow$ & \textbf{0.097} & \textbf{0.182} & \underline{0.114} & \textbf{0.022} & \textbf{0.104} & \textbf{0.093} & \textbf{0.291} & \underline{0.123} & \textbf{0.034} & \textbf{0.135} \\
\bottomrule
\end{tabular}}
\end{table*}

\begin{figure*}[t]
    \centering
    \includegraphics[width=1\textwidth]{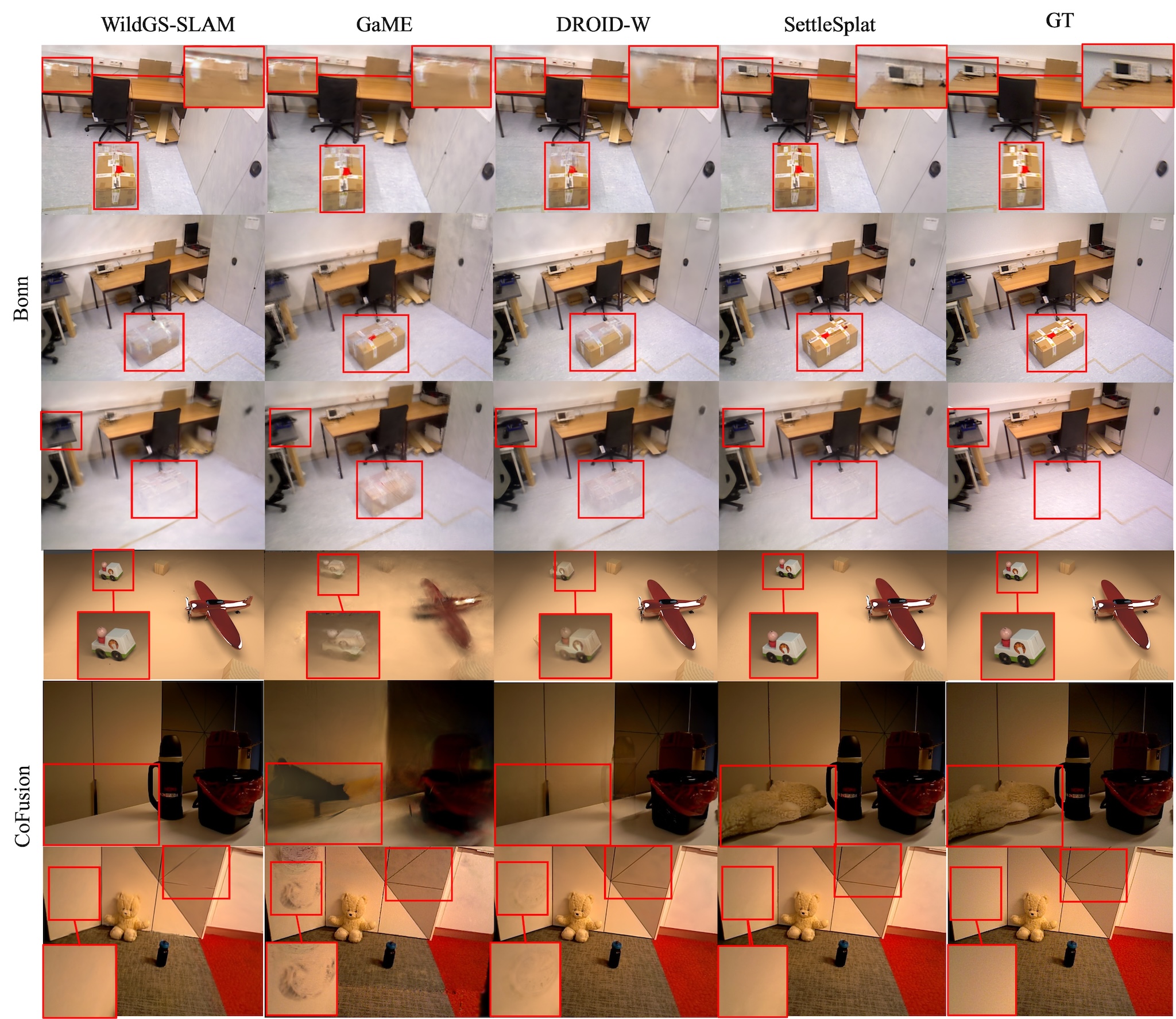}
    \caption{\textbf{Qualitative comparison} on Bonn (rows 1--3) and Co-Fusion (rows 4--6), covering moving, adding, and removing objects. Red boxes show the change regions. At old object locations, baselines leave ghosting artifacts, while SettleSplat reconstructs the updated scene cleanly. For added and moved objects, SettleSplat preserves appearance and detail, whereas the baselines blur or miss them. For removed objects, SettleSplat is competitive with other baselines while reducing ghosting.}
    \label{fig:pipeline}
\end{figure*}
\textbf{Quantitative Results.}
Table~1 and Table~2 report novel view synthesis on two datasets across 11 sequences. On Bonn~\cite{bonn}, SettleSplat achieves the best average PSNR, SSIM, and LPIPS on both the full image and the change region. 
The improvement is clearer in the change region on moving and placing sequences, where the scene structure changes and the map must be updated. Distractor-removal baselines~\cite{zheng2025wildgs,li2026droid} match or slightly exceed us on the removing sequences, but drop once an object is moved or added, since they remove observations rather than update the map. Our change-region performance remains close to full-image scores across all scenes, showing consistent map without ghost artifacts at the old object locations.
A similar trend is observed on Co-Fusion~\cite{cofusion} (Table~2), where we obtain the best average on both regions. On adding (\texttt{place}) and moving (\texttt{car}) sequences, we outperform all baselines, with change-region performance comparable to full-image scores. 
\blue{Notably, \texttt{place} sequence contains multiple sequential placements, where our method achieves the highest change-region PSNR, demonstrating its ability to accommodate multiple changes before final settlement.}
As on Bonn, distractor-removal baselines remain competitive on \texttt{sliding} and \texttt{handover}, where an object leaves the view and the map must delete rather than update Gaussians. 
GaME~\cite{yugay2026game} performs worse despite using sensor depth and instance masks, as its out-of-view change handling can fuse slow in-view manipulation into the static map, leaving artifacts (Figure~3).
\\
\textbf{Qualitative Results.}
Figure~3 demonstrates novel view synthesis on Bonn and Co-Fusion. 
At old object locations, our method shows the updated scene clearly, while the baselines leave ghosting artifacts. 
This happens because the old location is genuinely static in pre-change frames, so baselines map it as valid geometry and have no mechanism to detect that later frames contradict it.
For added objects, we preserve their appearance and detail, whereas slow body motion causes other methods to blur or miss them. For removal, SettleSplat is comparable to distractor-removal baselines~\cite{zheng2025wildgs,li2026droid} and avoids the residual ghosting left by long-term change methods~\cite{yugay2026game}.
\begin{table}[t]
\centering
\caption{Ablation study on CoFusion (averaged over all sequences). \textbf{Best} results are highlighted.}
\label{tab:ablation}
\resizebox{\linewidth}{!}{%
\begin{tabular}{l ccc ccc}
\toprule
 & \multicolumn{3}{c}{Full Image} & \multicolumn{3}{c}{Change Region} \\
\cmidrule(lr){2-4} \cmidrule(lr){5-7}
Variant & PSNR$\uparrow$ & SSIM$\uparrow$ & LPIPS$\downarrow$ & PSNR$\uparrow$ & SSIM$\uparrow$ & LPIPS$\downarrow$ \\
\midrule
w/o $V_d$ & 29.57 & 0.907 & 0.140 & 27.00 & 0.877 & 0.221 \\
w/o $K$ & 27.56 & 0.888 & 0.161 & 24.50 & 0.839 & 0.259 \\
w/o $M$ & {32.79} & {0.921} & {0.106} & {32.84} & {0.920} & {0.139} \\
Full (Ours) & \textbf{33.84} & \textbf{0.924} & \textbf{0.104} & \textbf{33.12} & \textbf{0.923} & \textbf{0.135} \\
\bottomrule
\end{tabular}}
\end{table}

\begin{table}[t]
\centering
\caption{Runtime comparison with dynamic-scene methods. We report the tracking speed in frames per second and the average mapping time per optimization iteration. Best results are in \textbf{bold}.}
\label{tab:runtime}
\resizebox{0.85\linewidth}{!}{%
\begin{tabular}{l cc cc}
\toprule
 & \multicolumn{2}{c}{Bonn} & \multicolumn{2}{c}{CoFusion} \\
\cmidrule(lr){2-3} \cmidrule(lr){4-5}
Method & \makecell{Track.\\FPS$\uparrow$} & \makecell{Map.\\/iter(s)$\downarrow$} & \makecell{Track.\\FPS$\uparrow$} & \makecell{Map.\\/iter(s)$\downarrow$} \\
\midrule
WildGS-SLAM~\cite{zheng2025wildgs} & 1.7 & 14.7 & 4.0 & 14.0 \\
DROID-W~\cite{li2026droid} & \textbf{10.7} & \textbf{1.3} & 20.4 & \textbf{1.2} \\
\textbf{SettleSplat (Ours)} & 10.6 & 1.8 & \textbf{21.7} & 1.4 \\
\bottomrule
\end{tabular}}
\end{table}
\subsection{Ablation Study}
Table 3 reports the ablation on the main components. In \textit{w/o} $V_d$, we disable the rendered depth uncertainty and use the raw depth residual as the geometry contradiction, which degrades performance. In \textit{w/o} $K$, we disable settlement detection and apply $W=r^2$ to all frames. 
Just-settled objects then remain suppressed by the lagging tracking uncertainty, so the map never catches up to the change.
In \textit{w/o} $M$, we mask dynamics in 2D pixel space instead of lifting them into 3D, and remove the correction of wrongly suppressed regions. The full model outperforms all variants on both regions, showing the effectiveness of each component.
Table~4 reports runtime. We exclude GaME here, as it performs mapping only and relies on segmentation precomputed offline. Compared with DROID-W, our method adds a small overhead from rendering depth uncertainty and computing the reliability weight.
For tracking accuracy, SettleSplat matches DROID-W and WildGS-SLAM, showing that our mapping stage does not degrade performance. Full results are in the supplementary.
\section{Conclusion}
In this paper, we introduce in-view scene changes, a challenging setting where transient and persistent dynamics are entangled within the camera's view. 
To address it, we propose a monocular mapping stage that turns tracking uncertainty into 3D dynamicity field and detects geometric contradictions.
Experiments on Bonn and Co-Fusion show better rendering quality on full images and in change regions, while keeping tracking and mapping speed comparable to the backbone.
By separating transient distractors from persistent changes, SettleSplat can maintain clean, up-to-date Gaussian maps from monocular RGB input.
\\
\textbf{Limitations.}
Our distractor handling relies on DINOv2 inconsistency, which can misfire and remove a static object instead of mapping it. Adding verification steps, such as segmentation priors, could improve these cases.



{
    \small
    \bibliographystyle{ieeenat_fullname}
    \bibliography{main}
}
\clearpage
\setcounter{page}{1}
\setcounter{section}{0}
\setcounter{table}{0}
\setcounter{figure}{0}

\maketitlesupplementary

\makeatletter
\renewcommand \thesection{S\@arabic\c@section}
\renewcommand\thetable{S\@arabic\c@table}
\renewcommand \thefigure{S\@arabic\c@figure}
\makeatother

\begin{table}[h]
\centering
\small
\setlength{\tabcolsep}{4pt}
\renewcommand{\arraystretch}{1.15}
\caption{SettleSplat hyperparameters used in the Bonn and Co-Fusion experiments.}
\label{tab:livesplat_hyperparams}
\begin{tabular}{@{}ccl@{}}
\toprule
\textbf{Symbol} & \textbf{Value} & \textbf{Description} \\
\midrule
\multicolumn{3}{@{}l}{\textit{Dynamicity pruning}} \\
$\tau_\delta$  & $0.5$  & Prune Gaussians with dynamicity \\
$\tau_n$       & $20$   & Min.\ observations before a Gaussian is prunable \\
$\epsilon$       & $1e-6$   & Geometry contradiction computation \\
\midrule
\multicolumn{3}{@{}l}{\textit{Mapping loss weights}} \\
$\lambda_\text{pho}$          & $0.5$ & Photometric (RGB) vs.\ depth loss weight\\
$\lambda_{\text{SSIM}}$ & $0.2$ & SSIM weight within the photometric loss \\
$\lambda_\text{iso}$ & $10$ & Scale regularizer \\
$\lambda_\text{dyn}$       & $0.3$ & Loss weight for per-Gaussian dynamicity \\
\midrule
\multicolumn{3}{@{}l}{\textit{Settlement frame ($K$) estimation}} \\
$\omega$                 & $5$    & Moving-average window over keyframes \\
$\tau_{\mathrm{settle}}$ & $0.15$ & Relative settle threshold for $K$ \\
\bottomrule
\end{tabular}
\end{table}

\begin{table}[h]
\centering
\small
\setlength{\tabcolsep}{5pt}
\renewcommand{\arraystretch}{1.15}
\caption{Held-out novel-view frames per scene.}
\label{tab:heldout}
\begin{tabular}{@{}llrrr@{}}
\toprule
\textbf{Dataset} & \textbf{Scene} & \textbf{$t_{\mathrm{hold}}$} & \textbf{End} & \textbf{\#} \\
\midrule
\multirow{7}{*}{Bonn}
 & moving\_nonobstructing\_box   & 476 & 776 & 61 \\
 & moving\_nonobstructing\_box2  & 703 & 933 & 47 \\
 & placing\_nonobstructing\_box  & 346 & 716 & 75 \\
 & placing\_nonobstructing\_box2 & 394 & 674 & 57 \\
 & placing\_nonobstructing\_box3 & 491 & 661 & 35 \\
 & removing\_nonobstructing\_box  & 291 & 491 & 41 \\
 & removing\_nonobstructing\_box2 & 647 & 902 & 52 \\
\midrule
\multirow{4}{*}{Co-Fusion}
 & place-items    & 910 & 970 & 21 \\
 & sliding-clock  & 600 & 753 & 52 \\
 & teddy-handover & 480 & 525 & 16 \\
 & car4-noise     & 367 & 400 & 12 \\
\bottomrule
\end{tabular}
\end{table}

In Supplementary Material, we provide more details to support the conclusions in the main paper.
\begin{itemize}
    \item \textbf{Implementation.} Hardware and software details and corresponding hyperparameters used in the main results.
    \item \textbf{Evaluation protocol.} More details about dataset preparation and evaluation protocol.
    \item \textbf{Additional results.} Per-scene tracking results on Bonn RGB-D Dynamic dataset.
\end{itemize}

\paragraph{System details.} We implement SettleSplat using Python 3.8, Pytorch 2.1.0 with CUDA 11.8 on NVIDIA H100 80 GB GPU. Starting from 3DGS rasterizer with pose \cite{kerbl3Dgaussians,monogs}, we render per-pixel depth uncertainty map by updating renderer with Equation 4 and 5 in the main paper.

\paragraph{Hyperparameters.} We follow our backbone \cite{li2026droid} for fair comparison. 
We report the settings used in our experiments with detailed descriptions in Table S1. The hyperparameters are same for all scenes on both Bonn RGB-D and CoFusion datasets.

\paragraph{Benchmark.} To obtain novel-view test frames, we annotate the first held-out frame $t_\text{hold}$. From $t_\text{hold}$ to the end of sequence, every 5th frame in Bonn and 3rd frame in CoFusion are withheld from keyframing/mapping and used only for novel-view synthesis and evaluation. 
A fixed interval avoids selecting favorable test views and keeps the split reproducible. The interval is denser on CoFusion because its sequences are shorter, ensuring enough novel views for stable evaluation, as shown in Table S2.
Notably, settlement frame $K$ is computed from the mapping frames without using held-out views. The specific $t_\text{hold}$ frame for each scene is described in Table S2. In Figure S4, we manually choose change regions with bounding boxes based on held-out test images. Then, these bounding boxes are applied across all methods to ensure fair comparisons.

\begin{figure*}[t]
    \centering
    \includegraphics[width=1\textwidth]{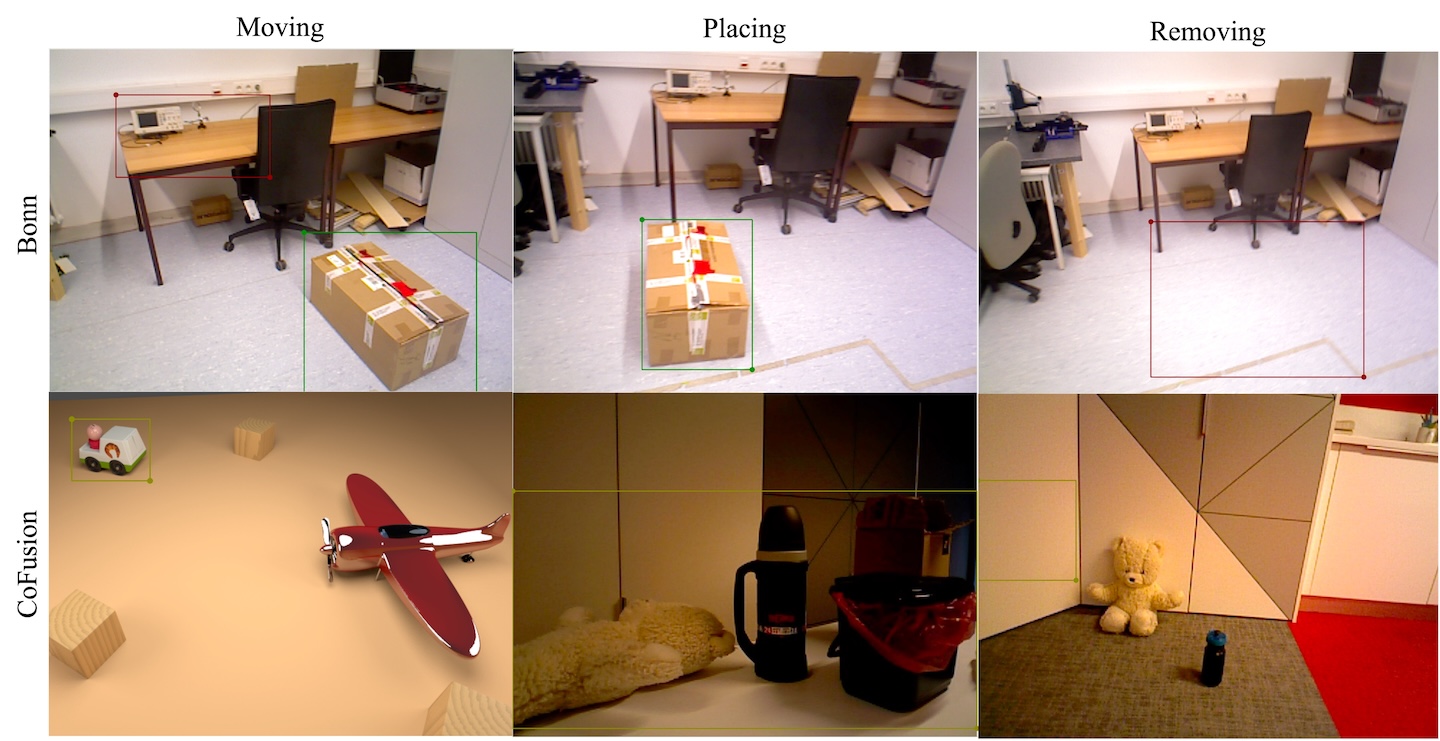}
    \caption{\textbf{Data preparation} with change region (bounding boxes) on Bonn and CoFusion, covering moving, adding and removing objects. The rendering metrics are computed on full rendered images and cropped change regions.}
    \label{fig:pipeline}
\end{figure*}

\paragraph{Additional Results.} Table S3 shows tracking performance with current baselines on Bonn RGB-D dynamic scenes \cite{bonn}, while CoFusion \cite{cofusion} ground truth pose is not available. SettleSplat maintains comparable tracking accuracy (Table S3) and speed (main Table 4), showing our mapping pipeline can be plugged in the uncertainty-aware tracking backbone without hurting its performance.

\begin{table}[t]
\centering
\setlength{\tabcolsep}{5pt}
\renewcommand{\arraystretch}{1.1}
\caption{Per-scene tracking accuracy (ATE RMSE [cm]) on Bonn dynamic scenes.}
\label{tab:ate_bonn}
\resizebox{\linewidth}{!}{%
\begin{tabular}{@{}lccc@{}}
\toprule
Scene & WildGS-SLAM \cite{zheng2025wildgs} & DROID-W \cite{li2026droid} & SettleSplat (Ours) \\
\midrule
\texttt{moving}    & 1.59 & 1.61 & 1.61 \\
\texttt{moving2}   & 2.61 & 2.38 & 2.38 \\
\texttt{placing}   & 1.52 & 1.61 & 1.58 \\
\texttt{placing2}  & 1.86 & 1.84 & 1.84 \\
\texttt{placing3}  & 1.99 & 2.04 & 2.08 \\
\texttt{removing}  & 1.40 & 1.49 & 1.49 \\
\texttt{removing2} & 1.99 & 2.02 & 2.01 \\
\midrule
Avg       & 1.85 & 1.86 & 1.86 \\
\bottomrule
\end{tabular}}
\end{table}


\end{document}